\documentclass[times,twocolumn, final]{elsarticle}
\usepackage{geometry}
\usepackage{framed,multirow}
\usepackage{amsmath}
\usepackage{amssymb}
\usepackage{latexsym}
\usepackage{subfigure}
\usepackage{graphicx}
\usepackage{booktabs}
\usepackage{threeparttable}
\usepackage{marginnote}
\usepackage{stfloats}
\usepackage{url}
\usepackage{xcolor}
\usepackage{multirow}
\definecolor{newcolor}{rgb}{.8,.349,.1}
\usepackage{array}

\usepackage{soul}
\usepackage{color, xcolor}
\soulregister{\cite}7

\usepackage{hyperref}

\usepackage[switch,pagewise]{lineno} 

\journal{Computers \& Graphics}

\begin{document}


\begin{frontmatter}

\title{Transferable Class Statistics and Multi-scale Feature Approximation for 3D Object Detection}%

\tnotetext[tnote1]{This work is supported by Natural Science Research Start-up Foundation of Recruiting Talents of Nanjing University of Posts and Telecommunications (NY224038)}

\author[1]{Hao Peng\corref{cor1}}
\cortext[cor1]{Corresponding author. penghao@njupt.edu.cn
}
\author[2]{Hong Sang}
\author[1]{Yajing Ma}
\author[1]{Ping Qiu}
\author[3]{Chao Ji}
\address[1]{School of Internet of Things, Nanjing University of Posts and Telecommunications, Nanjing 210003, China.}
\address[2]{College of Marine Electrical Engineering, Dalian Maritime University, Dalian 116026, China.}
\address[3]{College of Information Science and Engineering, Northeastern University, Shenyang 110819, China.}


\begin{abstract}
This paper investigates multi-scale feature approximation and transferable features for object detection from point clouds.
Multi-scale features are critical for object detection from point clouds. 
However, multi-scale feature learning usually involves multiple neighborhood searches and scale-aware layers, which can hinder efforts to achieve lightweight models and may not be conducive to research constrained by limited computational resources.
This paper approximates point-based multi-scale features from a single neighborhood based on knowledge distillation.  
To compensate for the loss of constructive diversity in a single neighborhood, this paper designs a transferable feature embedding mechanism. 
Specifically, class-aware statistics are employed as transferable features given the small computational cost. 
In addition, this paper introduces the central weighted intersection over union for localization to alleviate the misalignment brought by the center offset in optimization. 
Note that the method presented in this paper saves computational costs.
Extensive experiments on public datasets demonstrate the effectiveness of the proposed method.
\textit{The code will be released at https://github.com/blindopen/TSM-Det-Pointcloud-}
\end{abstract}


\end{frontmatter}



\section{Introduction}

The detection of 3D objects from point clouds is crucial for computer vision and scene understanding\cite{R1, R2, R3, R4, R5}, achieving the classification and localization of objects in the scene. 
Multi-stage voxel-based 3D object detection\cite{DENG2021VOXELRCNN, VoxSet} has gained more attention due to its higher detection accuracy. However, the required training resources, such as GPUs and large multimodal datasets, are also gradually increasing. Single-stage point-based 3D object detection methods\cite{PENG2024CPC3DET} can achieve lighter network models but receive relatively less attention from researchers. This paper aims to strike a balance between detection accuracy and model lightness in the single-stage point-based method.

Multi-scale feature learning in 3D object detection is usually required to meet the diversity of objects.
The local receptive field towards the voxel-based\cite{ZHENG2021CIASSD, ZHENG2021SESSD, YANG2023GDMAE, XIAO2023BSAODet} convolutional neural networks (CNN) makes the physical perception diverse. 
However, the quantization losses brought by voxelization are not conducive to capturing geometric features. 
Multi-scale grouping (MSG) proposed by PointNet++\cite{CHARLES2017POINTNET++} is widely utilized for feature aggregation\cite{SHI2019POINTRCNN, YANG20203DSSD,	ZHANG2022NOT, CHEN2022SASA, PENG2024CPC3DET}. 
However, MSG requires searching for points in multi-scale neighborhoods.
Moreover, the feature aggregation network at each scale is usually not shared.

\begin{figure}[!t]
	\centerline{\includegraphics[width=\columnwidth]{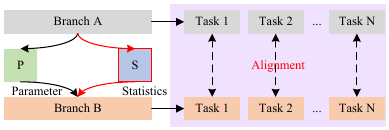}}
	\caption{Transferable Class Statistics and Multi-scale Feature Approximation based on knowledge distillation.}
	\label{Figure_diagram}
\end{figure}

Multi-neighborhood search and non-shared network parameters increase the computational cost, posing obstacles for researchers with limited resources.
Existing point-based feature aggregation methods\cite{ZHANG2022NOT, CHEN2022SASA, PENG2024CPC3DET} do not account for the mining of multi-scale neighborhood features within a single neighborhood.
This paper approximates point-based multi-scale features from a single neighborhood based on knowledge distillation. 
As shown in Fig. \ref{Figure_diagram}, the teacher branch A with multi-scale neighborhood search is first pre-trained. Then, this paper constructs the student branch B with a single-scale neighborhood search to approximate the performance of A.
Note that A does not participate in the inference of the model.
The decrease in neighborhood searches reduces the computational cost.
The task alignments constrain the consistency of feature learning between branches.
The main challenge of single-neighborhood multi-scale feature learning is the information loss caused by the reduction of neighborhood search.
This paper explores the knowledge distillation and multi-task alignment mechanism in single-neighborhood multi-scale learning to promote multi-scale feature approximation.

Reducing some neighborhood searches and the corresponding feature aggregation layers makes the model lightweight.
However, the challenge of feature learning emerged.
Specifically, reducing parameters and simplifying the network structure have somewhat reduced the feature learning capability.
Plug-and-play modules are generally effective in improving network performance with minimal computational overhead, but they often overlook the transferability of feature construction.
To drive feature learning in the student branch, this paper incorporates class-aware statistics\cite{PENG2024CPC3DET}, as demonstrated in Fig. \ref{Figure_diagram}.
In this paper, class-aware statistics are characterized as a tensor of dimensions $N \times D$, with $N$ signifying the number of object categories for detection.
The class-aware statistics generated by the teacher branch are integrated into the classification head.
Moreover, the prediction layers embedding class-aware statistics are utilized to guide the feature learning of MLPs (Multi-layer Perceptrons) for localization.
This paper enhances the feature learning performance with minimal additional computational cost, as the class-aware statistics are merely vectors of tensor type.

The center voting unifies the feature aggregation and promotes the performance of single-stage point-based 3D object detection.
The potential hotspot center plays a significant role in various aspects of single-stage architecture, including object-level feature integration and optimization alignment based on Intersection over Union (IoU).
This paper adopts center-weighted IoU (CwIoU)\cite{PENG2024CPC3DET} to alleviate the misalignment due to center offsets in knowledge distillation when fitting the teacher performance by the student branch.

\begin{figure*}[!t]
	\centerline{\includegraphics[width=\textwidth]{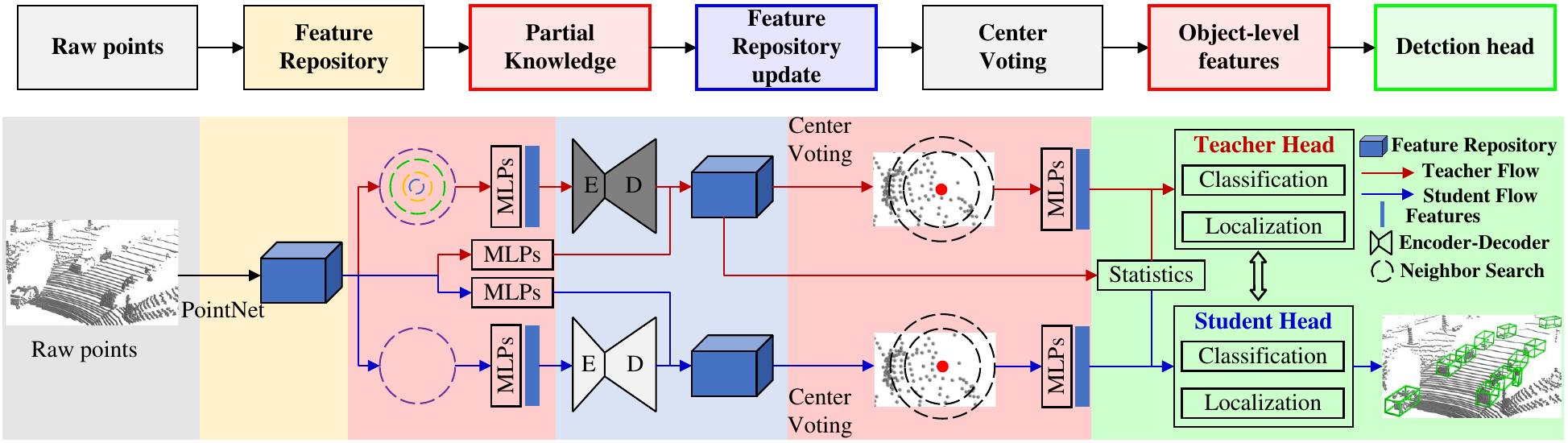}}
	\caption{{The architecture of TSM-Det. Utilizing raw point data with PointNet, a feature repository is developed. To enhance this repository, knowledge from high-confidence foreground points is extracted. The potential object hotpot is obtained by center voting. Object-level features are captured and then combined with the statistics-embedded detection head to achieve object detection.}}
	\label{Figure_architecture}
\end{figure*}
This paper innovatively utilizes single-neighborhood point clouds to approximate multi-scale features based on knowledge distillation for point-based 3D object detection, and explores its key issues and challenges. In addition, this paper further studies the application of class-aware statistics in object detection model compression. 
The works mentioned above are effective for model lightweighting and have not yet been applied in other object detectors. 
Moreover, we also consider the misalignment caused by the center offsets between teacher and student predictions.
Compared to sparse-supervised or semi-supervised learning\cite{r2p1,r2p2,r2p3,r2p4}, TSM-Det prioritizes balancing computational cost and detection accuracy during inference.

In short, this paper explores multi-scale feature approximation and transferable features for single-stage point-based 3D object detection. The essential contributions can be outlined as follows.

\begin{enumerate}
	\item Multi-scale features are approximated from a single neighborhood based on knowledge distillation, balancing model performance and complexity.
	\item Transferable class-aware statistics are introduced to drive feature learning and enhance detection performance. 
	\item To tackle the misalignment caused by center offsets between IoU and optimization in knowledge distillation, center-weighted IoU loss is employed.
	\item The innovative detection method TSM-Det introduced in this paper has demonstrated competitive performance on public datasets.
\end{enumerate}

The related works are presented in Section II, followed by a
detailed explanation of our proposed 3D object detector, TSM-
Det, in Section III. Section IV gives the extensive experimental results, while Section V discusses the work and its limitations. Section VI concludes the paper.
	
\section{Related Work}
\subsection{Object Detection Architecture for Point Clouds}
Single-stage 3D object detectors\cite{LANG2019POINTPILLARS} achieve lightweight and save computation costs.	SASSD\cite{HE2020SASSD} uses attachable segmentation and center prediction layers to aid discriminative feature learning.
SE-SSD\cite{ZHENG2021SESSD} leverages knowledge distillation to enhance the effectiveness of CIA-SSD\cite{ZHENG2021CIASSD}.
In contrast, two-stage point cloud object detection usually lays more emphasis on the design of the refinement modules.
RoI feature representation based on sparse tensors\cite{YAN2018SECOND,DENG2021VOXELRCNN},  graph\cite{QIAN2022BANET},  and attention mechanism\cite{SHENG2021CT3D, HU2022PDV} has received wide attention.
In addition, 3D object detection can be divided into voxel-based\cite{DENG2021VOXELRCNN}, point-based\cite{CHARLES2017POINTNET++, ZHANG2022NOT, CHEN2022SASA}, and voxel-and-point-based\cite{SHI2020PVRCNN, HE2022SVGANET} methods according to the point cloud representations.
This paper focuses on the single-stage voxel-based 3D object detection method and introduces a novel detector TSM-Det.
\subsection{Multi-scale Features Learning from Point Clouds}
Multi-scale feature aggregation is critical for discriminative feature learning. Voxels-based convolution\cite{LANG2019POINTPILLARS, DENG2021VOXELRCNN, ZHENG2021SESSD} and neighborhood queries\cite{YANG20203DSSD, CHEN2022SASA, ZHANG2022NOT} are popularly employed in multi-scale feature construction from point clouds.
Multiple neighborhood queries joint symmetric function is typically used to achieve multi-scale feature aggregation in point-based methods\cite{CHARLES2017POINTNET++, PENG2024CPC3DET}. 
However, points within a single neighborhood typically present challenges in multi-scale feature learning.
In this paper, we approximate multi-scale features based on feature learning areas in a single neighborhood.
\subsection{Lightweight Feature Enhancement from Point Clouds}
PV-RCNN\cite{SHI2020PVRCNN} transfers sparse features to key points and then learns the geometric features based on the key points to aid the refinement.
The point-based method downsamples the points in the scene and selects represent points to accelerate inference. 
F-FPS, S-FPS, and confidence-based filtering are usually employed for point cloud downsampling\cite{YANG20203DSSD, CHEN2022SASA, PENG2024CPC3DET, ZHANG2022NOT}.
Voxel R-CNN\cite{DENG2021VOXELRCNN} implements neighborhood search based on sparse tensors, accelerating traditional PointNet in point cloud feature learning.
Methods based on projection and Range images\cite{LIANG2021RANGEIOUDET} represent another option for lightweight networks, given that they transfer 3D space data to 2D pseudo-image space, saving computing costs.
The attention mechanism\cite{SHENG2021CT3D,MAHMOUD2022DVF,MAO2021VOTR,VoxSet} captures contextual relationships and enhances feature discriminability.
However, the above methods rarely consider the transfer of feature information from the training to the inference.

\section{Proposed Method}
\label{sec:Proposed Method}
This paper explores multi-scale feature approximation and transferable class statistics based on the single-stage 3D object detector.
As shown in Fig. \ref{Figure_architecture}, knowledge distillation is introduced into the multi-scale feature approximation, and network compression is also achieved.
The class-aware statistics generated in the teacher branch are transferred to the detection head to enhance performance. 
Moreover, the auxiliary detection head and the CwIoU loss are employed for knowledge distillation.
Ultimately,  we presented TSM-Det, a lightweight detector with a single-stage architecture.
\subsection{Feature Repository Initialization}
\label{sec:Feature Repository Initialization}
\begin{figure}[!ht]
	\centerline{\includegraphics[width=\columnwidth]{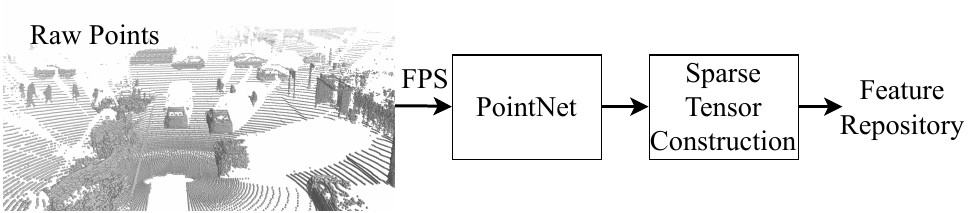}}
	\caption{Flow of feature repository initialization}
	\label{Figure_rp2fr}
\end{figure}
The feature repository is a sparse representation of the point cloud scene. 
As shown in Fig.\ref{Figure_rp2fr}, this paper first constructs the feature repository based on the raw point clouds.
Define the point cloud scene as $\textbf{P}=\left\lbrace p_i = \left( x_i,y_i,z_i,r_i\right)  \right\rbrace_{i=1 \ldots N}$, where $\left( x,y,z\right) $represents the coordinates of the $i$th point, and $r$ denotes its reflection intensity.
Farthest Point Sampling (FPS) is employed to select representative points
$\mathbf{P_r}=\left\lbrace p_i = \left( x_i,y_i,z_i,r_i\right)  \right\rbrace_{i=1 \ldots M}$
to reduce the computational cost. Then, local features $\mathbf{F_r}$ of representative points are learned based on raw points and PointNet\cite{CHARLES2017POINTNET++}.
This paper builds a sparse tensor based on $\mathbf{P_r}$, where the grid size is $\left[ \Delta w,\Delta l,\Delta h\right] $.
The mean of coordinates and features $\left\lbrace \left( vx_i,vy_i,vz_i\right), f_i \right\rbrace_{i=1 \ldots K}$ are calculated to characterize the grid with numerous points.
Finally, the initial feature repository can be expressed algebraically as $\textbf{R}=\left\lbrace v_i = \left( vx_i,vy_i,vz_i\right), f_i \right\rbrace_{i=1 \ldots K}$.
Note that usually $N > M \geq K$ here.

\subsection{Multi-scale Feature Approximation}
\label{sec:Multi-scale Feature Approximation}

\begin{figure}[!ht]
	\centerline{\includegraphics[width=\columnwidth]{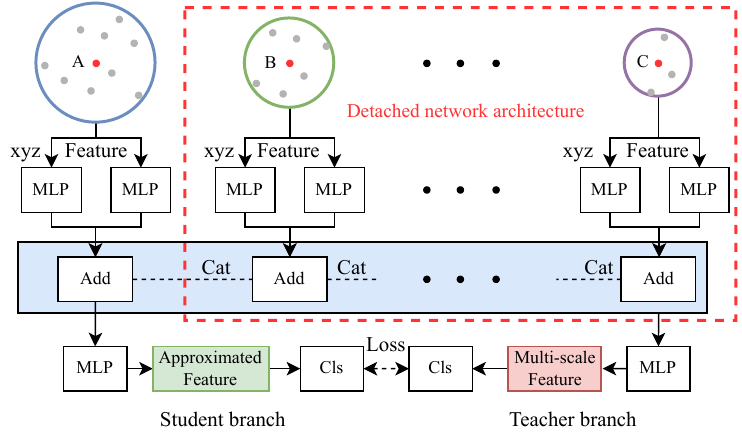}}
	\caption{Multi-scale features approximation in single-scale neighborhood}
	\label{Figure_MFS}
\end{figure}
The feature repository $\textbf{R}$ realizes the characterization of key points of the scene. Based on the feature repository, this paper further mines point cloud features and lightweights the detection model, as shown in Fig. \ref{Figure_architecture}.
This Section proposes a multi-scale feature approximation aimed at mining local multi-scale features within a single-scale neighborhood.
Feature learning leverages neighborhood multi-scale search, which is more straightforward for creating lightweight models compared to those based on local receptive field convolution\cite{LIU2024MSRMNet}.

As shown in Fig.\ref{Figure_MFS}, the features mined in the single-scale neighborhood should perform similarly to multi-scale features. This paper introduces the classification for feature alignment.
The multi-scale feature branch and single-scale neighborhood branch are employed as the teacher and student, respectively. 
The teacher branch provides soft labels to supervise the student parameter learning for classification.
This paper selects some representative points using S-FPS\cite{CHEN2022SASA} and approximates the multi-scale features of the teacher branch through single-scale neighborhoods in the feature repository.
The features approximated using partial representative points are referred to as partial knowledge.
As shown in Fig.\ref{Figure_MFS}, single-neighborhood multi-scale feature learning employs the point cloud of a single scale(A) and the corresponding network architecture.
The number of neighbor searches in the inference of TSM-Det is reduced from $N$ to 1, where $N$ is the number of scales in the teacher branch.
Additionally, the detached network architecture reduces computation costs.

Partial knowledge, i.e., the approximated multi-scale features, is utilized to update the feature repository.
Parameter redundancy is commonly observed in the design of deep neural networks.
This paper performs parameter compression for updating the repository. 
Define partial knowledge as $\textbf{K}=\left\lbrace p_i = \left( x_i,y_i,z_i\right), k_i  \right\rbrace_{i=1 \ldots J}$, where $\left( x,y,z\right) $ and $k$ are coordinates and knowledge features respectively. 
First, a sparse tensor $\mathbf{K_R}$ is constructed using spatial mapping, mirroring the structure of the feature repository.
Note that zeros are assigned for sparse voxels with no features. 
Furthermore, an encoder-decoder is designed based on sparse convolution to mine scene features. 
Finally, the scene features processed by MLPs are added to the features of the feature repository.
Supposing the algebraic expression of encoder-decoder is $ED$, and the feature repository foreground confidence is $\mathbf{S_R}$.
The feature repository is updated as Eq. \ref{eq1}.
\begin{equation}
\label{eq1}
	\mathbf{R} = \mathbf{S_R} \cdot ED\left( \mathbf{K_R}\right)  + MLPs\left( \mathbf{R}\right) 
\end{equation}
\begin{figure}[!ht]
	\centerline{\includegraphics[width=\columnwidth]{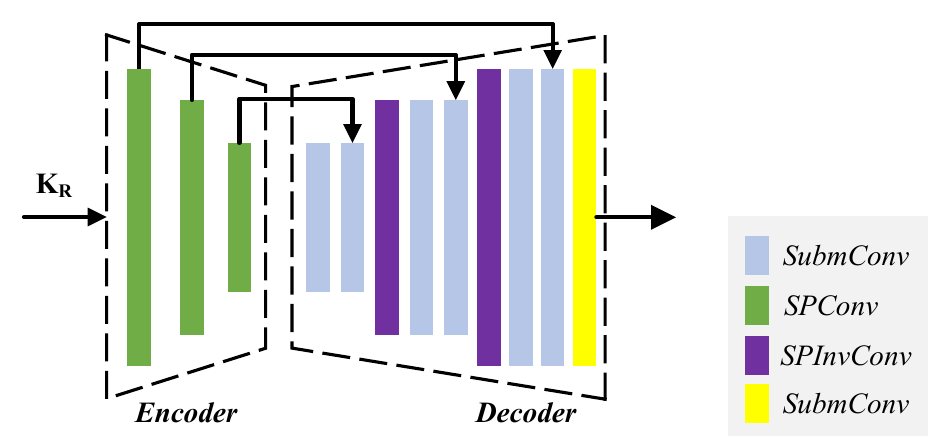}}
	\caption{The network structure of the encoder-decoder}
	\label{Figure_ed}
\end{figure}

This paper employs the encoder-decoder in CPC-3Det\cite{PENG2024CPC3DET} as the teacher and the lightweight network as the student to achieve parameter compression.
The structure of the encoder-decoder in this paper is shown in Fig. \ref{Figure_ed}.
Lightweight sparse convolution layer and shortcut are used to implement scene feature learning.

\subsection{Object-level Feature Aggregation}
Center voting is usually used in point-based 3D object detection to sense the potential location of objects. 
Object-level feature aggregation aims to obtain potential hotspot representations.
In this paper, partial knowledge is utilized to predict the center through the Linear layer. In addition, the object-level features are aggregated based on the feature repository and predicted centers.
This paper introduces class-aware statistics from the teacher branch (detailed in Section III.D). 
Therefore, the same number of neighborhood searches as in the teacher branch is retained here. 
Note that only half of the neighborhood points in the teacher branch are utilized for feature aggregation in the student branch.

\subsection{Transferable Class-aware Statistics}
\begin{figure}[!ht]
	\centerline{\includegraphics[width=.7\columnwidth]{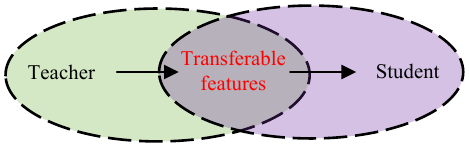}}
	\caption{Transferable features for teacher and student branches.}
	\label{Figure_transfer}
\end{figure}

This paper introduces transferable features to enhance the feature extraction process in the student branch.
As shown in Fig. \ref{Figure_transfer}, transferable features are derived from the teacher, given that the teacher branch usually has better feature learning abilities than the student branch. 
Furthermore, the transferable features will be embedded in or guide the student branch to enhance detection performance.
The class-aware statistics\cite{PENG2024CPC3DET} are applied as transferable features in this paper.
Specifically, the similarity between object-level and class-aware features is useful for classification.
In addition, the parameter learning of the localization branch can be participated in by class-aware features.

\begin{figure}[!ht]
	\centerline{\includegraphics[width=\columnwidth]{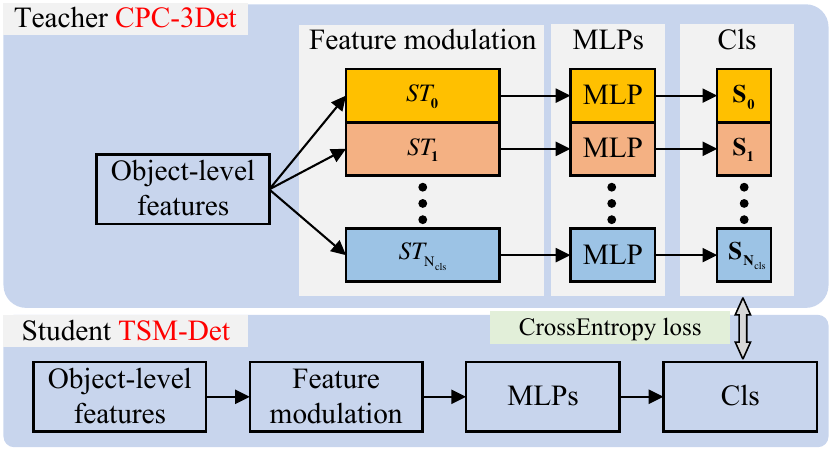}}
	\caption{The teacher and student branches for classification}
	\label{Figure_CLS}
\end{figure}

As shown in Fig. \ref{Figure_CLS}, the teacher branch utilizes class-aware statistics to modulate object-level features and employs MLPs to predict category confidence scores. 
Specifically, the multiplication operation is used to achieve feature modulation here.
TSM-Det proposed in this paper maintains the same network architecture as the teacher branch. 
Note that CPC-3Det serves as a teacher to guide the parameter learning of TSM-Det.
\begin{figure}[!ht]
	\centerline{\includegraphics[width=\columnwidth]{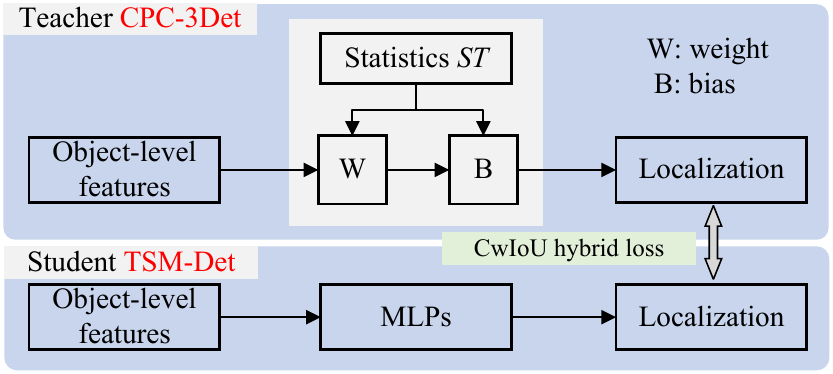}}
	\caption{The teacher and student branches for localization}
	\label{Figure_Loc}
\end{figure}

For localization in CPC-3Det, class-aware statistics participate in the weights and biases learning. However, this method consumes more computing resources than utilizing MLPs directly. 
As depicted in Fig. \ref{Figure_Loc}, with the integration of knowledge distillation, TSM-Det allows MLPs to match the performance of the localization head embedded with class-aware statistics. 
In addition,  CwIoU hybrid loss\cite{PENG2024CPC3DET} is used to quantify the offset between teacher and student.

\subsection{Training Losses}
TSM-Det employs soft and hard loss to jointly drive the optimization of parameters, as shown in Eq. \ref{eq2}.
\begin{equation}
\label{eq2}
	L = \lambda_{soft} \cdot L_{soft}+ \lambda_{hard} \cdot L_{hard}
\end{equation}

Here $L_{soft}$ and $L_{hard}$ represent soft loss and hard loss, respectively.
$\lambda_{soft}$ and $\lambda_{hard}$ are the weights for soft loss and hard loss, respectively.
Here, we empirically set $\lambda_{soft}=0.7$ and $\lambda_{hard}=0.3$.
\subsubsection{Soft Loss}
Suppose that the confidence of the non-normalized category is $C=\left\lbrace c_1,c_2, \cdots, c_n\right\rbrace $, where $n$ refers to the count of classes to be detected. 
The temperature parameter $T$ and the $Sigmoid$ layer are utilized to normalize the confidence, and the algebraic expression is Eq. \ref{eq3}
\begin{equation}
\label{eq3}
	C_{norm} = \frac{1}{\left( 1+ \exp \left( -C/T \right)  \right) }
\end{equation}

Where $C_{norm}$ is the normalized result of the predicted confidence. The temperature parameter of the soft label in the classification is $T=3$.
Furthermore,  Focal loss is used to quantify the difference between predicted confidence and soft labels in classification.
As shown in Eq. \ref{eq4}, $C_{norm}^{soft}$ and $C_{norm}^{pred}$ are the soft labels and the predicted values calculated by Eq. \ref{eq3}, respectively.
We set $\alpha_{t}=0.25$ and $\gamma=2$ here.
\begin{equation}
\label{eq4}
	\begin{split}
		&{L}_{soft}^{cls}=-\alpha_{t}\left(C_{norm}^{soft}-p_{t}\right)^{\gamma} \log \left(p_{t}\right) \\
		&\text { where } p_{t}=\left\{\begin{array}{ll}
			C_{norm}^{pred} & \text { for foreground class} \\
			1-C_{norm}^{pred} & \text { otherwise }
		\end{array}\right.
	\end{split}
\end{equation}

Define the 3D bounding boxes predicted by the student and the soft label output by the teacher as $Bbox^{pred}$ and $Bbox^{soft}$, respectively. 
The center coordinates, length, width, and height of the bounding box are $\left( x,y,z\right) $, $w$, $l$, and $h$, respectively. Then the algebraic representation of the soft loss function for localization is provided in Eq. \ref{eq5}.
Note that only positive samples, i.e., foreground sample points, are considered in Eq. \ref{eq5}. The regression layer initializes its parameters using the Kaiming method. Since the teacher branch is pre-trained, it consistently matches the ground truth. Moreover, the student branch prediction remains constrained by hard labels.
\begin{equation}
\label{eq5}
	\begin{split}
		L_{soft}^{loc} &= IoULoss\left( Bbox^{pred}, Bbox^{soft}\right)\\
		&+\lambda_{ind}\cdot\sum{smoothL1({{[\Delta b]}})}_{{b\in \left( x,y,z,w,l,h \right)}}\\
		&+\lambda_{corner}\cdot L_{corner}
	\end{split}	
\end{equation}
\subsubsection{Hard Loss}
TSM-Det proposed in this paper is a supervised method. Hard labels are provided to supervise the network parameter learning. 
We adopt cross-entropy loss in the classification of the detection head as well as the auxiliary branches. 
For localization, this paper quantifies the difference between the output and the ground truth through IoU loss, corner regularization term, and SmoothL1 loss.

\section{Experiments}
To adapt to different datasets, the network hyperparameter configurations vary accordingly. For the evaluation of the $val$ split and $test$ set in KITTI, the network structure and hyperparameters of the detector are the same, with only differences in training samples. The ablation experiment in this manuscript is only conducted on the KITTI $val$ split due to the substantial size of the Waymo dataset.
\subsection{Datasets}
\subsubsection{KITTI Dataset}
The TSM-Det proposed in this paper is verified on the KITTI dataset\cite{KITTI}, which provides 7481 training samples and 7518 test samples for 3D object detection from laser point clouds. The training samples are divided into a $train$ split with 3712 samples and a $val$ split with 3769 samples. 
This paper focuses on the performance of TSM-Det in the $val$ split and the $test$ set.
Additionally, the model that is evaluated on the $test$ set is trained on the combined $train$ and $val$ splits.
\subsubsection{Waymo Open Dataset}
TSM-Det is also evaluated on the Waymo Open dataset\cite{WAYMO}. 
TSM-Det is trained by 20\% of data randomly selected from 798 training scenes (158361 sweeps) and evaluated by the whole 202 validation scenes (40077 sweeps). 
TSM-Det is capable of detecting vehicles, pedestrians, and cyclists.

\begin{table*}[!t]
	\centering
	\caption{Evaluation of TSM-Det on The Test Set of KITTI}
	\label{table7.1}
    \small
	\begin{tabular}{p{80pt}<{\centering}p{20pt}<{\centering}p{22pt}<{\centering}p{20pt}<{\centering}p{20pt}<{\centering}p{22pt}<{\centering}p{20pt}<{\centering}p{20pt}<{\centering}p{22pt}<{\centering}p{20pt}<{\centering}p{20pt}<{\centering}p{22pt}<{\centering}p{20pt}<{\centering}}
		\hline
		\multirow{2}{*}{Method} &
		\multicolumn{3}{c}{Car (3D Detection)}&
		\multicolumn{3}{c}{Car (BEV Detection)}&
		\multicolumn{3}{c}{Cyclist (3D Detection)}&
		\multicolumn{3}{c}{Cyclist (BEV Detection)}\cr\cline{2-13}   
		&Easy&Mod.&Hard&Easy&Mod.&Hard&Easy&Mod.&Hard&Easy&Mod.&Hard\\
		\hline
		\multicolumn{13}{c}{Two-stage}\cr\cline{1-13}
				Part-$A^{2}$\cite{partA2}& 87.81& 78.49& 73.51&91.70 &	87.79 &	84.61&79.17&63.52&56.93&83.43&68.73&61.85\\
		PV-RCNN\cite{SHI2020PVRCNN}& 90.25& 81.43& 76.82&94.98&90.65&86.14&78.60&63.71&57.65&82.49&68.89&62.41\\	
		Voxel R-CNN\cite{DENG2021VOXELRCNN}& 90.90& 81.62& 77.06&94.85&88.83&86.13&-&-&-&-&-&-\\
	VoxSet\cite{VoxSet}&88.53&82.06&77.46&92.70&89.07&86.29&-&-&-&-&-&-\\
		\hline
		\multicolumn{13}{c}{Single-stage}\cr\cline{1-13}
		PointPillars\cite{LANG2019POINTPILLARS}& 82.58& 74.31& 68.99&90.07&86.56&82.81&77.10&58.65&51.92&79.90&62.73&55.58\\
		3DSSD\cite{YANG20203DSSD}&88.36& 79.57 &74.55&92.66&89.02&85.86&\textbf{82.48}&\textcolor{blue}{\underline{{64.10}}}&\textcolor{blue}{\underline{{56.90}}}&\textbf{85.04}&\textcolor{blue}{\underline{{67.62}}}&61.14\\
		SASSD\cite{HE2020SASSD}&88.75&79.79& 74.16&\textbf{95.03}&\textbf{91.03}&{85.96}&-&-&-&-&-&- \\
		CIA-SSD\cite{ZHENG2021CIASSD}&\textbf{89.59} & 80.28& 72.87&93.74&89.84&82.39&-&-&-&-&-&-\\	
		SASA\cite{CHEN2022SASA}& 88.76 &\textcolor{blue}{\underline{{82.16}}}&\textcolor{blue}{\underline{77.16}}&92.87&89.51&\textcolor{blue}{\underline{{86.35}}}&-&-&-&-&-&- \\
		SVGA-Net\cite{HE2022SVGANET}& 87.33 &80.47 &75.91&92.07&\textcolor{blue}{\underline{{89.88}}}&85.59&78.58&62.28&54.88&81.25&66.82&59.37\\
		IA-SSD\cite{ZHANG2022NOT}& 88.34 & 80.13& 75.04&92.79 &89.33&84.35 &\textcolor{blue}{\underline{{82.36}}}&\textbf{66.25} &\textbf{59.70}&81.30&66.29&59.58\\
		GD-MAE\cite{YANG2023GDMAE}& 88.14&79.03 &73.55&\textcolor{blue}{\underline{{94.22}}}&88.82&83.54&-&-&-&-&-&-\\		
		{CPC-3Det\cite{PENG2024CPC3DET}}& 88.80&\textbf{82.27}&\textbf{77.24}&93.00&89.61&84.55&78.58&63.56&56.68&{82.59}&{67.61}&\textcolor{blue}{\underline{{61.42}}}\\
		\textbf{TSM-Det (Ours)}& \textcolor{blue}{\underline{{88.93}}}&82.15&77.04&92.99&89.62&\textbf{86.51}&80.15&63.71&56.66&\textcolor{blue}{\underline{{84.48}}}&\textbf{68.24}&\textbf{61.82}\\
		\hline
		\multicolumn{13}{p{250pt}}{(Mod.) Moderate.}
	\end{tabular}
\end{table*}

\subsection{Implementation}
\subsubsection{Network}
Regarding the KITTI dataset, this work considers the scene corresponding to the forward image, analyzing point clouds within the specified ranges: $\left[ 0m,70.4m\right] $ for the $X$-axis, $\left[ -40m,40m\right] $ for the $Y$-axis, and $\left[ -3m,1m\right]$ for the $Z$-axis.
In this work, CPC-3Det with default parameters is employed as the teacher branch.
For evaluation on the KITTI, the student branch aggregates key point neighborhood features based on PointNet to initialize the feature repository. 
4096 key points are utilized, incorporating multi-scale information across radius ranges of $\left[0.2, 0.4, 0.8\right] m$.
The corresponding network channel configurations are $[16, 16, 32]$, $[16, 16, 32]$, and $[32, 32, 64]$.
The final aggregated feature consists of 64 channels.
The count of partial points is 512.
Moreover, this paper only considers 32 points within a radius of $1.6m$. The dimension of the MLPs here is $[128, 256, 512]$.
The specified parameters of the encoder-decoder in the student are $[64, 64, 128, 64, 64]$.
The feature dimension in the updated repository is 128.
Only 16 points and corresponding features at each scale are considered in the object-level feature mining.
The student branch directly introduces class-aware statistics from the teacher.
In addition, TSM-Det does not utilize the statistic-embedded localization head but employs a convolutional neural network with a 1x1 kernel size for localization.
For the Waymo Open dataset, TSM-Det initializes the repository with 8192 feature points, while utilizing 1024 partial points.

\subsubsection{Data Augmentation}
The GT (Ground Truth) sampling\cite{YAN2018SECOND} is introduced to data augmentation. 
For the KITTI dataset, scenes are flipped along the $X$-axis randomly. The random rotation uniform sampled in [-Pi/4, Pi/4] and scaling uniform sampled in [0.9, 1.1] are utilized to enhance the diversity of scenes.
In addition, the object-level random noise is introduced to enrich the diversity. 
Please note that flipping along the $Y$-axis is also taken into account for the Waymo Open dataset.
\subsubsection{Platform and Model Parameters}
The Adam optimizer and the learning rate with the one\_cycle strategy are employed to drive the parameters learning.
The learning rate is set to 0.01 initially. 
The parameters of the teacher network are frozen during training.
We train our TSM-Det on an NVIDIA 2080Ti GPU.
For the KITTI dataset, 16384 points are fed as the training data for 100 epochs, and the batch size is set to 16. 
20000 points are retained for inference. 
For the Waymo Open dataset, 65534 points are preserved as the training data for 30 epochs, and the batch size is set to 8. During the inference, 163820 points are maintained for each scene.
\subsection{Evaluation Implemented On The KITTI Dataset}
\subsubsection{KITTI test set}
We compare TSM-Det with recent SOTA 3D detection methods.
As shown in Table \ref{table7.1}, black bold font indicates the best performance, while blue underline demonstrates the second best performance.
TSM-Det demonstrates superior performance using fewer parameters when compared to CPC-3Det.
For moderate and hard Car 3D object detection, TSM-Det is only slightly lower than CPC-3Det 0.12\% and 0.2\% mAP, respectively.
Compared with CPC-3Det, the TSM-Det proposed in this paper improved by 1.96\% mAP in the BEV detection of hard Cars and achieved better results overall for Cyclist detection.
Voxel R-CNN\cite{DENG2021VOXELRCNN} and VoxSet\cite{VoxSet} are two-stage 3D object detectors
that tend to achieve higher detection accuracy. It cannot be ignored that Voxel R-CNN requires a large number of anchors to be preset. It is worth noting that the proposed method TSM-Det is superior than the two-stage method VoxSet.
\subsubsection{KITTI val split}
\begin{table}[!th]
	\centering
	\caption{Evaluation of TSM-Det on The Val Split of KITTI}
	\label{table2} 
	\begin{tabular}{p{80pt}<{\centering}p{30pt}<{\centering}p{22pt}<{\centering}p{22pt}<{\centering}p{22pt}<{\centering}}
		\hline
		\multirow{2}{*}{Method} &
		\multirow{2}{*}{RP}&
		\multicolumn{3}{c}{Car (3D Detection)}\cr\cline{3-5}  
		& & Easy&Mod.&Hard\\
		\hline
		\multicolumn{5}{c}{Two-stage}\cr\cline{1-5}
		PV-RCNN\cite{SHI2020PVRCNN}&11 & 89.35& 83.69& 78.70\\
		Voxel R-CNN\cite{DENG2021VOXELRCNN}&11 & 89.41& 84.52& 78.93\\		
		$\rm H^{2}3$D R-CNN\cite{DENG2021H23DRCNN}&11& 89.63 & 85.20&\textcolor{blue}{\underline{79.08}}\\
		BADet\cite{QIAN2022BANET}&11&\textcolor{blue}{\underline{90.06}}&\textcolor{blue}{\underline{85.77}}&79.00 \\
		SPG\cite{XU2021SPG}&40& 92.53& 85.31& 82.82\\	
		PV-RCNN\cite{SHI2020PVRCNN}&40&{92.57}& 84.83& 82.69\\
		Voxel R-CNN\cite{DENG2021VOXELRCNN}&40& 92.38& 85.29& 82.86\\
		EQ-PVRCNN\cite{YANG2022EQPVRCNN}&40& 92.52& 85.61&\underline{83.13}\\
		PDV\cite{HU2022PDV}&40& 92.56& 85.29& 83.05\\
		BSAODet\cite{XIAO2023BSAODet}&40&92.27&85.06&82.75\\
		\hline
		\multicolumn{5}{c}{Single-stage}\cr\cline{1-5}	
		3DSSD\cite{YANG20203DSSD}&11&89.71&79.45&78.67\\
		SASSD\cite{HE2020SASSD}&11&\textcolor{blue}{\textbf{90.15}}&79.91& 78.78 \\	
		CIA-SSD\cite{ZHENG2021CIASSD}&11 & 90.04& 79.81& 78.80\\
		SVGA-Net\cite{HE2022SVGANET}&11&90.59&80.23&\textcolor{blue}{\textbf{79.15}}\\	
		{CPC-3Det\cite{PENG2024CPC3DET}}&11& 89.58& \textcolor{blue}{\textbf{86.09}}&78.94\\
		SASA\cite{CHEN2022SASA}&40& 92.19 &\underline{85.76}& 83.11 \\
		GD-MAE\cite{YANG2023GDMAE}&40& -&82.01 &-\\
		{CPC-3Det\cite{PENG2024CPC3DET}}&40& \textbf{92.86}& \textbf{86.02}& \textbf{83.29}\\
		\cline{1-5}  
		
		\textbf{TSM-Det (Ours)}&11&89.56& 85.69&78.83\\
		
		\textbf{TSM-Det (Ours)}&40&\underline{92.73}& 85.62&82.99\\
		\hline
		\multicolumn{5}{p{200pt}}{(RP) Recall Positions, (Mod.) Moderate.}\\
	\end{tabular}
\end{table}

We evaluate the performance of TSM-Net on the KITTI $val$ split.
The best results for mAP@40 and mAP@11, calculated from 40 and 11 recall positions, are highlighted in black bold and blue bold, respectively.
The suboptimal result is underlined.
Table \ref{table2} shows that TSM-Net has attained competitive results in mAP@11 and mAP@40 relative to CPC-3Det.
TSM-Net outperforms the single-stage methods presented in Table II regarding mAP@11 for moderate cars.
Compared to 3DSSD\cite{YANG20203DSSD}, our method has achieved mAP@11 improvements of 6.24\% and 0.16\%, respectively, with moderate and hard difficulty.
TSM-Det outperformed  Voxel R-CNN\cite{DENG2021VOXELRCNN} and $\rm H^{2}3$D R-CNN\cite{DENG2021H23DRCNN} with improvements of 1.17\% and 0.49\% mAP@11 in the moderate cars.
Furthermore, TSM-Det also earned good performance in mAP@40.
TSM-Det is better than SPG\cite{XU2021SPG}, PV-RCNN\cite{SHI2020PVRCNN}, Voxel R-CNN\cite{DENG2021VOXELRCNN}, and PDV\cite{HU2022PDV} in mAP@40 of the moderate car by 0.31\%, 0.79\%, 0.33\%, and 0.33\%  respectively.
With lighter parameters than SASA\cite{CHEN2022SASA}, TSM-Det performs somewhat lower at moderate and hard difficulty levels but demonstrates good performance in easy car detection.

\begin{table}[!h]
	\centering
	\caption{Evaluation of TSM-Det for BEV Detection on the Val of KITTI}
	\label{table3}
	\small
	\begin{tabular}{p{75pt}<{\centering}p{30pt}<{\centering}p{30pt}<{\centering}p{30pt}}
		\hline
		\multirow{2}{*}{Method} &
		\multicolumn{3}{c}{Car (BEV Detection) mAP@40}\cr\cline{2-4}  
		& Easy&Moderate&Hard\\
		\hline
		PV-RCNN\cite{SHI2020PVRCNN}& 95.76& 91.11& 88.93\\
		Voxel R-CNN\cite{DENG2021VOXELRCNN}& 95.52& 91.25& 88.99\\
		SPG\cite{XU2021SPG}& 94.99& 91.11& 88.86\\
		PDV\cite{HU2022PDV}& 93.60& 91.14& \textbf{90.74}\\
		BADet\cite{QIAN2022BANET}&90.63&88.86&88.10\\
		{CPC-3Det\cite{PENG2024CPC3DET}}&96.02&\textbf{91.94}&89.41\\
		\textbf{TSM-Det (Ours)}&\textbf{96.05}&91.85&89.37\\
		\hline
	\end{tabular}
\end{table}

We further evaluate TSM-Det on BEV car detection using the KITTI validation split, as indicated in Table \ref{table3}.
Competitive performance is obtained between TSM-Det and CPC-3Det. 
Compared with other 3D detection methods listed in Table \ref{table3}, TSM-Det exhibits significant strengths at easy and moderate difficulty levels.
As shown in Table \ref{table4}, TSM-Det also achieves comparable performance to CPC-3Det in cyclists. 
Note that the performance of TSM-Det in the pedestrian is not as good as that of CPC-3Det. 
Pedestrian requires features to be more discriminable to overcome the noise caused by similar objects.
\begin{table}[!t]
	\centering
	\caption{Evaluation of TSM-Det for Pedestrian and Cyclist in KITTI Val}
	\label{table4}
	\small
	\begin{tabular}{p{10pt}<{\centering}p{70pt}<{\centering}p{12pt}<{\centering}p{12pt}<{\centering}p{12pt}<{\centering}p{12pt}<{\centering}p{12pt}<{\centering}p{12pt}<{\centering}}
		\hline
		\multirow{2}{*}{Item} &
		\multirow{2}{*}{Method}&
		\multicolumn{3}{c}{Ped. mAP@40}&
		\multicolumn{3}{c}{Cyc. mAP@40}\cr\cline{3-8}  
		& &Easy&Mod.&Hard&Easy&Mod.&Hard\\
		\hline
		\multirow{2}{*}{3D}
		&CT3D\cite{SHENG2021CT3D}&65.73&58.56&53.04&\textbf{91.99}&71.60&67.34\cr
		&DVF+PV\cite{MAHMOUD2022DVF}&66.08& 59.18& 54.68&90.93& 72.46&68.05\cr
		&SE-SSD\cite{ZHENG2021SESSD}&63.27& 57.32& 50.82&80.07& 70.43&66.45\cr
		&{CPC-3Det\cite{PENG2024CPC3DET}}&\textbf{66.22}&\textbf{60.27}&\textbf{55.25}&89.39&\textbf{72.84}&\textbf{68.40}\cr
		\cline{2-8}
		&\textbf{TSM-Det (Ours)}&63.03&57.92&52.26&91.63&72.13&67.48\cr
		\hline
		\multirow{2}{*}{BEV}
		&SE-SSD\cite{ZHENG2021SESSD}&67.47&61.88&55.94&{91.83}&72.62&68.24\cr
		&{CPC-3Det\cite{PENG2024CPC3DET}}&\textbf{68.61}&\textbf{64.25}&\textbf{58.38}&91.12&{75.07}&\textbf{70.63}\cr
		\cline{2-8}
		&\textbf{TSM-Det (Ours)}&66.47&61.68&56.76&\textbf{92.70}&\textbf{75.20}&70.54\cr
		\hline
		\multicolumn{8}{p{200pt}}{(Ped.) Pedestrian, (Cyc.) Cyclist, (Mod.) Moderate.}\\
	\end{tabular}
\end{table}
\begin{figure*}[!t]
	\centerline{\includegraphics[width=\textwidth]{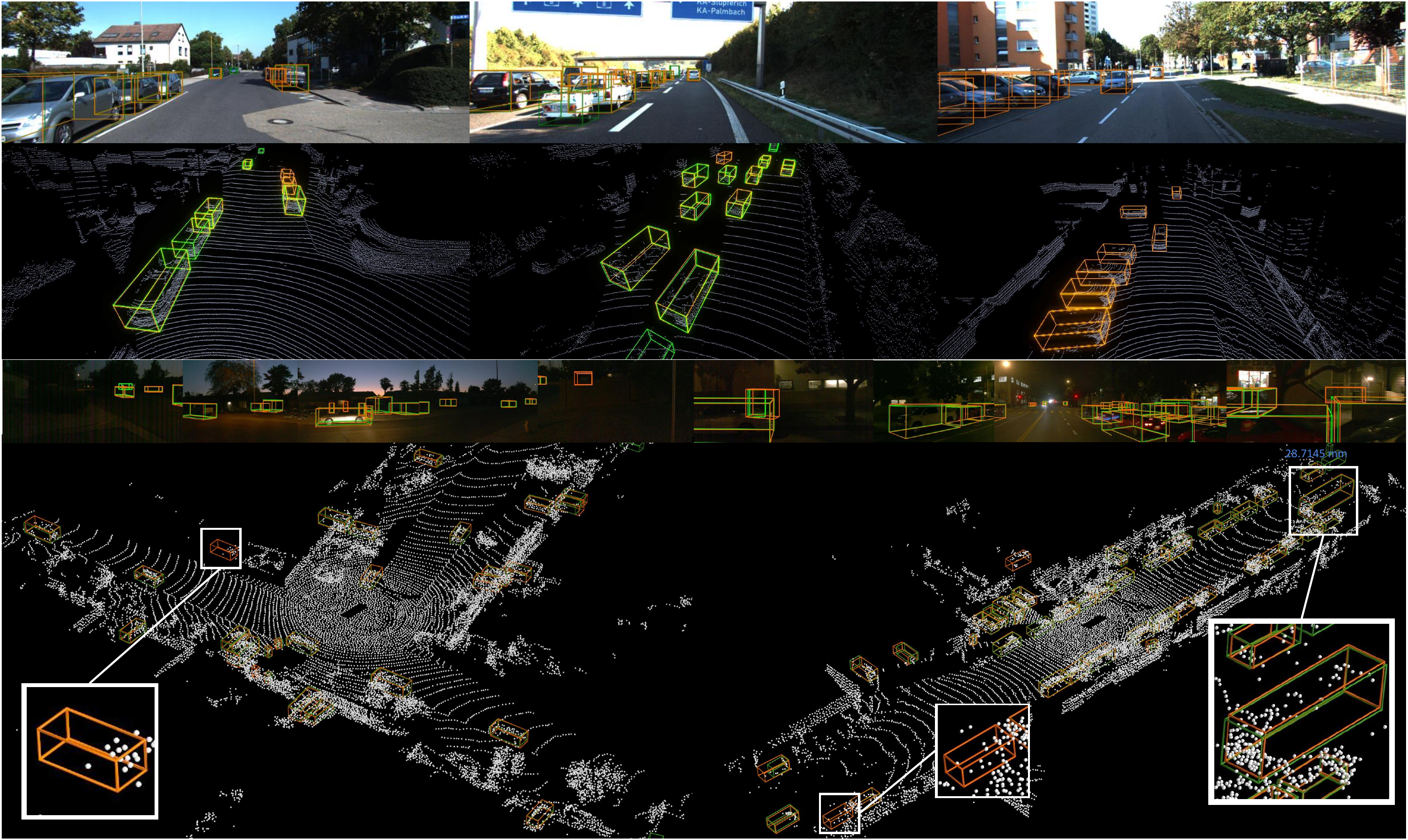}}
	\caption{Displayed in the top two rows are the qualitative results for the KITTI dataset, while the last two rows showcase the Waymo Open dataset. Ground truth 3D bounding boxes are marked in green,  while the results predicted by TSM-Det are indicated in orange.}
	\label{Figure_qualitative}
\end{figure*}
\begin{table*}[!t]
	\centering
	\caption{Results of TSM-Det on the Waymo Open dataset}
	\label{table5}
	\small
	\begin{tabular}{p{120pt}<{\centering}p{18pt}<{\centering}p{21pt}<{\centering}p{18pt}<{\centering}p{21pt}<{\centering}p{18pt}<{\centering}p{21pt}<{\centering}p{18pt}<{\centering}p{21pt}<{\centering}p{18pt}<{\centering}p{21pt}<{\centering}p{18pt}<{\centering}p{21pt}<{\centering}}
		\hline
		\multirow{2}{*}{Method} &
		\multicolumn{2}{c}{Veh. (L\_1)}&
		\multicolumn{2}{c}{Veh. (L\_2)}&
		\multicolumn{2}{c}{Ped. (L\_1)}&
		\multicolumn{2}{c}{Ped. (L\_2)}&
		\multicolumn{2}{c}{Cyc. (L\_1)}&
		\multicolumn{2}{c}{Cyc. (L\_2)}\cr\cline{2-13}  
		&mAP&mAPH&mAP&mAPH&mAP&mAPH&mAP&mAPH&mAP&mAPH&mAP&mAPH\\
		\hline
		\multicolumn{13}{c}{100\%Trainset}\cr\cline{1-13}
		SECOND+DSA\cite{MAO2021DSAPVRCNN}&71.10& 70.70& 63.40& 63.00&-&-&-&-&-&-&-&-\\
		IA-SSD\cite{ZHANG2022NOT}&70.53&69.67&61.55&60.80&69.38&58.47&60.30&50.73&67.67&65.30&64.98&62.71\\
		\hline
		\multicolumn{13}{c}{20\%Trainset}\cr\cline{1-13}
		PointPillars\cite{LANG2019POINTPILLARS}&70.43&69.83&62.18&61.64&66.21&46.32&58.18&40.64&55.26&51.75&53.18&49.80\\
		CenterPoint+P\cite{YIN2021CENTERNET}&70.50&69.96&62.18&61.69&\textbf{73.11}&\textbf{61.97}&\textbf{65.06}&\textbf{55.00}&65.44&63.85&62.98&61.46\\
		X-Ray SECOND-Scaled\cite{gambashidze2024weaktostrong}&68.30&64.30&61.90&58.00&-&-&-&-&-&-&-&-\\
		
		{CPC-3Det\cite{PENG2024CPC3DET}}&\textbf{72.14}&\textbf{71.61}&\textbf{63.49}&\textbf{63.02}&64.75&55.75&56.44&48.46&\textbf{68.00}&\textbf{66.10}&\textbf{65.41}&\textbf{63.58}\\
		\textbf{TSM-Det  (Ours)}&71.29&70.68&62.67&62.13&63.14 &53.73&55.10&46.74&64.71&62.61&62.24&60.22\\
		\hline
		\multicolumn{13}{p{400pt}}{(Veh.) Vehicle, (Ped.) Pedestrian,  (Cyc.) Cyclist, (L\_1) LEVEL\_1, (L\_2) LEVEL\_2, (-) Not available.}\\
	\end{tabular}
\end{table*}

\subsection{Evaluation Implemented on the Waymo Open Dataset}
This paper presents an evaluation of the TSM-Det on the Waymo Open dataset with 202 validation sequences. For Vehicle detection, TSM-Det outperforms PointPillars\cite{LANG2019POINTPILLARS}, and CenterPoint+P\cite{YIN2021CENTERNET}, as shown in Table \ref{table5}. In addition, TSM-Det performs better in vehicle detection than IA-SSD\cite{ZHANG2022NOT}, which uses all training samples.

\subsection{Qualitative analysis} 
The TSM-Det proposed in this paper intuitively achieves good detection results for car detection in the KITTI dataset and Waymo Open dataset, as Fig. \ref{Figure_qualitative} shows. 
Some large-sized and unlabeled objects are still detected. 
TSM-Det intuitively performs better on the KITTI dataset, while the Waymo Open dataset is more challenging due to its greater sample diversity and wider range of scenes.

\begin{table*}[!t]
	\centering
	\caption{Performance of Knowledge Distillation Framework in TSM-Det}
	\label{kd}
	\small
	\begin{tabular}{p{30pt}<{\centering}p{100pt}<{\centering}p{20pt}<{\centering}p{20pt}<{\centering}p{20pt}<{\centering}p{20pt}<{\centering}p{20pt}<{\centering}p{20pt}<{\centering}p{20pt}<{\centering}p{20pt}<{\centering}p{20pt}<{\centering}}
		\hline
		\multirow{2}{*}{Item} &
		\multirow{2}{*}{Method}&
		\multicolumn{3}{c}{Car mAP@40}&
		\multicolumn{3}{c}{Pedestrian mAP@40}&
		\multicolumn{3}{c}{Cyclist mAP@40}\cr\cline{3-11}  
		& &Easy&Mod.&Hard&Easy&Mod.&Hard&Easy&Mod.&Hard\\
		\hline
		\multirow{3}{*}{3D}
		&No knowledge distillation&92.32&83.72&80.94&59.85&53.50& 48.83&89.13&68.90&64.63\cr
		&\textbf{TSM-Det  (Ours)}&\textbf{92.73}& \textbf{85.62}&\textbf{82.99}&\textbf{63.03}&\textbf{57.92}&\textbf{52.26}&\textbf{91.63}&\textbf{72.13}&\textbf{67.48}\cr
		\cline{2-11}
		&improvement&+0.41&+1.9&+2.05&+3.18&+4.42&+3.43&+2.5&+3.32&+2.85\cr
		\hline
		\multirow{3}{*}{BEV}
		&No knowledge distillation&95.60& 89.72& 89.02&63.68&58.04& 52.66&90.22&71.38&67.29\cr
		&\textbf{TSM-Det (Ours)}&\textbf{96.05}&\textbf{91.85}&\textbf{89.37}&\textbf{66.47}&\textbf{61.68}&\textbf{56.76}&\textbf{92.70}&\textbf{75.20}&\textbf{70.54}\cr
		\cline{2-11}
		&improvement&+0.45& +2.13&+0.35&+2.79&+3.64&+4.1&+2.48&+3.82&+3.25\cr
		\hline
	\end{tabular}
\end{table*}

\subsection{Ablation Studies}
In order to be lightweight while ensuring accuracy in point cloud object detection, this paper uses knowledge distillation and hybrid loss during training. 
In fact, knowledge distillation and hybrid loss supervise multi-scale feature approximation, as well as classification and localization with the participation of transferable attention statistics. Additionally, this paper examines the performance of multi-scale feature approximation in classification and foreground segmentation. In addition, multi-category scenarios are also attempted with TSM-Det in this paper.

\subsubsection{knowledge distillation}
We only removed knowledge distillation, and the other configurations are consistent with TSM-Det.
As shown in Table \ref{kd}, the introduction of knowledge distillation has greatly improved the performance of the detector, especially for the 3D detection of moderate cars, pedestrians, and cyclists.
\subsubsection{Temperature}
\begin{figure}[!t]
	\centerline{\includegraphics[width=\columnwidth]{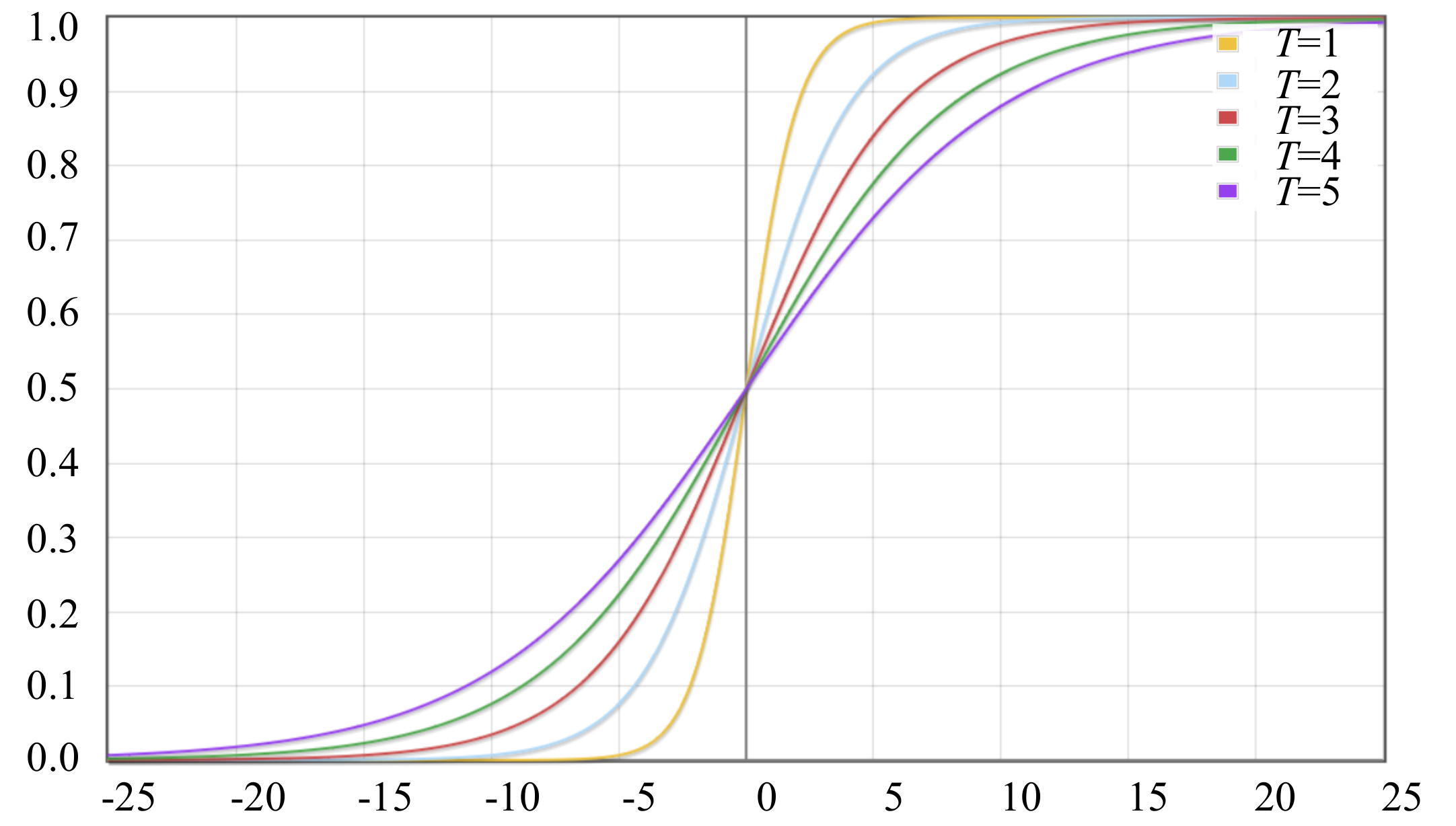}}
	\caption{Temperature control with different $T$ settings}
	\label{Figure_T}
\end{figure}
\begin{table}[!h]
	\centering
	\caption{Performance of different temperature parameters}
	\label{table7}
	\small
	\begin{tabular}{p{20pt}<{\centering}p{40pt}<{\centering}p{40pt}<{\centering}p{40pt}<{\centering}}
		\hline
		\multirow{2}{*}{$T$}&
		\multicolumn{3}{c}{Moderate (3D) mAP@40}\cr\cline{2-4}  
		& Car&Pedestrian&Cyclist\\
		\hline
		1 &\textbf{85.62}& \textbf{59.80}&71.03\\
		3 &\textbf{85.62}& 57.92&\textbf{72.13}\\
		5&85.54& 56.64&68.95\\
		\hline
	\end{tabular}
\end{table}
Temperature emphasizes the importance of negative samples in soft loss, as shown in Fig. \ref{Figure_T}.
The object detection performance of TSM-Det under different temperatures is shown in Table \ref{table7}.
The detection performance of TSM-Det decreases when $T$ is 5. We reason that blindly increasing $T$ reduces the discriminability between different class confidences.
The $T$ used in TSM-Det training in this paper is 3, which performs better in car and cyclist detection.
At the same time, we found that fixed parameters between objects cannot be applied to multiple categories when facing multi-object detection. Therefore, we will focus on adaptive parameter settings future.

\subsubsection{Multi-scale feature approximation}
\sethlcolor{yellow}
\begin{table}[!h]
	\centering
	\caption{The analysis of multi-scale feature approximation }
	\label{AMFA}
	\small
	\begin{tabular}{p{80pt}<{\centering}p{40pt}<{\centering}p{40pt}<{\centering}p{40pt}<{\centering}}

		\hline
		\multirow{2}{*}{Method}&
		\multicolumn{3}{c}{Moderate (3D) mAP@40}\cr\cline{2-4}  
		& Car&Pedestrian&Cyclist\\
            \hline
            \textbf{TSM-Det(Ours)} &\textbf{85.62}& \textbf{57.92}&\textbf{72.13}\\
		\hline
		no KD &83.72&53.50&68.90\\
            \emph{Improvement}(Ours)&\emph{+1.90}&\emph{+4.42}&\emph{+3.23}\\
            \hline
            no max range&85.46 &56.45 &67.65\\
            \emph{Improvement}(Ours)&\emph{+0.16}&\emph{+1.52}&\emph{+4.48}\\
            \hline
		
	\end{tabular}
\end{table}
The multi-scale feature approximation proposed in this paper is mainly based on knowledge distillation with classification alignment. Additionally, the neighborhood range will also impact the performance of the multi-scale feature approximation. As shown in Table \ref{AMFA}, knowledge distillation is crucial for fitting multi-scale features in a single neighborhood, improving 1.9\%, 4.42\%, and 3.23\% mAP for car, pedestrian, and cyclist, respectively. Note that if the neighborhood scale does not cover the maximum scale range in teacher branch, it will still have a negative impact on model performance.

We evaluate the effectiveness of our single-neighborhood multi-scale learning in point cloud classification and foreground segmentation tasks, on the Modelnet40 and KITTI datasets, respectively. This paper builds a segmentation dataset based on KITTI and considers the points in the box as foreground points.
\begin{table}[!h]
	\centering
	\caption{Multi-scale feature approximation in classification.}
	\label{mc}
	\small
	\begin{tabular}{p{80pt}<{\centering}p{70pt}<{\centering}p{70pt}<{\centering}}
		\hline
		{Method}&Instance Accuracy&Class Accuracy\cr
		\hline
		PointNet-MSG\cite{CHARLES2017POINTNET++}\dag& \textbf{93.10}&\textbf{91.58} \\
            \hline
            PointNet-S\cite{CHARLES2017POINTNET++}\dag&91.96 & 90.00\\
		\textbf{PointNet-MFA}&\textbf{92.73}& \textbf{90.48}\\
		\hline
		\multicolumn{3}{p{210pt}}{\dag Reproduced. (MFA) Multi-scale feature approximation.}
	\end{tabular}	
\end{table}

As shown in Table \ref{mc}, compared with the method that does not use multi-scale feature approximation, the proposed method improves classification performance and is closer to the feature effect achieved through multi-neighborhood aggregation. Moreover, the multi-scale feature approximation proposed in this paper also shows rationality in foreground point segmentation, as shown in Table \ref{mfs}.

\begin{table}[!h]
	\centering
	\caption{Multi-scale feature approximation in foreground segmentation.}
	\label{mfs}
	\small
	\begin{tabular}{p{120pt}<{\centering}p{100pt}<{\centering}}
		\hline
		{Method}&IoU\cr
		\hline
		PointNet-MSG\cite{CHARLES2017POINTNET++}\dag&66.34\\
            \hline
            PointNet-S\cite{CHARLES2017POINTNET++}\dag&65.87\\
		\textbf{PointNet-MFA}&\textbf{66.79}\\
		\hline
		\multicolumn{2}{p{210pt}}{\dag Reproduced. (MFA) Multi-scale feature approximation.}
	\end{tabular}	
\end{table}

\subsubsection{Transferable class-aware statistics}
\begin{table}[!h]
	\centering
	\caption{Performance@40 of the transferable class-aware statistics}
	\label{tcs40}
	\small	
	\begin{tabular}{p{40pt}<{\centering}p{40pt}<{\centering}p{40pt}<{\centering}p{40pt}<{\centering}}
		\hline
		\multirow{2}{*}{Embedded}&
		\multicolumn{3}{c}{Moderate (3D) mAP@40}\cr\cline{2-4}  
		& Car&Pedestrian&Cyclist\\
		\hline
		- &85.30&57.34&68.58\\
		\checkmark &\textbf{85.62}& \textbf{57.92}&\textbf{72.13}\\
		\hline
	\end{tabular}
\end{table}
We explore the effect of a classification detection head with transferable class-aware statistics embedding on TSM-Det. As shown in Table \ref{tcs40}, compared to conventional detection heads, the embedding transferable class-aware statistics promotes 3D object detection performance.
We believe that the transferable class-aware statistics utilize the valuable statistical knowledge in the teacher branch to promote the representation of the student branch and improve the detector reasoning performance. In addition, the class-aware statistics in TSM-Det are $ N \times 256$, where $N$ is the number of categories, resulting in only a slight computational cost. We particularly focus on the performance at the mean Average Precision at 11 points (mAP@11). As shown in Table \ref{tcs11}, incorporating transferable class-aware statistics significantly improves the model's overall performance, especially for cyclists.
\begin{table}[!h]
	\centering
	\caption{Performance@11 of the transferable class-aware statistics}
	\label{tcs11}
	\small	
	\begin{tabular}{p{20pt}<{\centering}p{40pt}<{\centering}p{30pt}<{\centering}p{30pt}<{\centering}p{30pt}<{\centering}}
		\hline
		\multirow{2}{*}{Class}&\multirow{2}{*}{Embeded}&
		\multicolumn{3}{c}{3D Detection mAP@11}\cr\cline{3-5}  
		&& Easy&Moderate&Hard\\
		\hline
		Car&- &\textbf{89.68}&84.79&78.68\\
		&\checkmark &{89.56}& \textbf{85.69}&\textbf{78.83}\\
		\hline
            Pedestrain&- &62.67&57.35&\textbf{53.87}\\
		&\checkmark &\textbf{63.58}& \textbf{58.58}&52.40\\
		\hline
            Cyclist&- &83.99&68.79&63.83\\
		&\checkmark &\textbf{87.55}& \textbf{71.12}&\textbf{66.10}\\
		\hline
	\end{tabular}
\end{table}

\subsubsection{Hybird soft losses}

The knowledge distillation provides soft labels for TSM-Det network training and drives network optimization through hybrid soft losses.
Specifically, the soft loss items are mainly related to classification, individual offset(Smooth L1 for centers, width, length, and height), CwIoU, and corner regularization.
To verify the contribution of each soft loss item, we designed ablation experiments.
The results in Table \ref{table9} show that classification, CwIoU, and corner regularization soft losses all promote the performance of TSM-Det, and the individual offset term also plays a positive role.
\begin{table}[!h]
	\centering
	\caption{Performance of the hybrid soft losses}
	\label{table9}
	\small
	\begin{tabular}{p{18pt}<{\centering}p{18pt}<{\centering}p{18pt}<{\centering}p{18pt}<{\centering}p{20pt}<{\centering}p{20pt}<{\centering}p{20pt}<{\centering}}
		\hline
		\multirow{2}{*}{Cls}&
		\multirow{2}{*}{offset}&
		\multirow{2}{*}{CwIoU}&
		\multirow{2}{*}{Corner}&
		\multicolumn{3}{c}{Moderate (3D) mAP@40}\cr\cline{5-7}  
		& & & &Car&Ped.&Cyclist\\
		\hline
		-&\checkmark&\checkmark &\checkmark &83.69& 56.32&71.36\\
		\checkmark &- &\checkmark &\checkmark& 85.47&\textbf{58.53}& 71.50\\
		\checkmark &\checkmark&-&\checkmark &83.56&57.92&70.56\\
		\checkmark &\checkmark&\checkmark&- &83.64&55.97&72.03\\
		\checkmark &\checkmark &\checkmark &\checkmark&\textbf{85.62}& 57.92&\textbf{72.13}\\
		\hline
		\multicolumn{6}{p{150pt}}{(Ped.) Pedestrian}\\
	\end{tabular}
\end{table}

We also examine the case where both the soft loss weight and the hard loss weight are set to 0.5. As shown in the Table\ref{tablesoftweight}, a slight loss of accuracy occurs when the soft loss weight is 0.5.
\begin{table}[!h]
	\centering
	\caption{Performance of the different soft loss weights}
	\label{tablesoftweight}
	\small
	\begin{tabular}{p{30pt}<{\centering}p{30pt}<{\centering}p{25pt}<{\centering}p{25pt}<{\centering}p{25pt}<{\centering}}
		\hline
		\multirow{2}{*}{Soft}&
		\multirow{2}{*}{Hard}&
		\multicolumn{3}{c}{Moderate {(3D)} mAP@40}\cr\cline{3-5}  
		& &Car&Ped.&Cyclist\\
		\hline
		0.5 & 0.5 &85.24& 57.72&71.71\\
		0.7 & 0.3& 85.62 & 57.92& 72.13 \\
		
		\hline
		\multicolumn{5}{p{100pt}}{(Ped.) Pedestrian}\\
	\end{tabular}
\end{table}

\subsection{Performance in scenes with complex objects}
\begin{table}[!h]
	\centering
	\caption{mAP performance in scenes with complex objects}
	\label{nuval}
	\small
	\begin{tabular}{p{25pt}<{\centering}p{30pt}<{\centering}p{30pt}<{\centering}p{30pt}<{\centering}p{30pt}<{\centering}p{30pt}<{\centering}}
		\hline
            \multirow{4}{*}{Teacher}
		& Car&truck&Cons.&bus&trailer\cr\cline{2-6}        
            &75.50&40.07 &14.09 &63.38&28.89\cr\cline{2-6}
		&barrier&motocycle&bicycle&pedestrain&traf.\cr\cline{2-6}
		&52.05 &41.88  & 21.45&75.16&46.63\\
		\hline
		\multirow{4}{*}{Student}
		& Car&truck&Cons.&bus&trailer\cr\cline{2-6}        
            &73.65&39.36 &13.87 &\textbf{63.41}&28.74\cr\cline{2-6}
		&barrier&motocycle&bicycle&pedestrain&traf.\cr\cline{2-6}
		&49.63 &41.23  & 20.80&73.07&42.80\\
		\hline
	\end{tabular}
\end{table}
This paper evaluates the performance of TSM-Det on the complex scene dataset nuScenes, which includes 10 categories of objects to be detected. The mAP and nuScenes detection score (NDS) are employed to evaluate the performance differences between the teacher branch and the student branch in this paper, as shown in Table \ref{nuval} and Table \ref{nuvalts} respectively.

\begin{table}[!h]
	\centering
	\caption{Performance of nuScenes detection score} 
	\label{nuvalts}
	\small
	\begin{tabular}{p{25pt}<{\centering}p{18pt}<{\centering}p{18pt}<{\centering}p{18pt}<{\centering}p{18pt}<{\centering}p{18pt}<{\centering}p{18pt}<{\centering}p{18pt}<{\centering}}
		\hline
		Method& mAP& mATE& mASE& mAOE& mAVE& AAE& NDS\\  
            \hline
            Teacher&45.91&28.08&24.34 &29.60&29.14&20.08&59.83\\
            \hline
		Student&44.66&\textbf{27.77}&\textbf{24.23}&\textbf{28.99}&29.17&20.25&59.29\\
		\hline
	\end{tabular}
\end{table}

For multi-category object detection, the student branch can still maintain strong consistency with the teacher branch and shows a slight advantage in bus detection, as demonstrated in Table \ref{nuval}. Additionally, the student branch exhibits performance improvements in mATE, mASE, and mAOE, as indicated in Table \ref{nuvalts}.
\subsection{Complexity analysis}
As shown in Table \ref{tablecomplex}, the lightweight point cloud object detection method TSM-Det proposed in this paper simplified CPC-3Det. 
Under a single 2080Ti graphics card, the maximum supported batch size of TSM-Det is twice that of CPC-3Det. 
Moreover, compared with other methods in Table \ref{tablecomplex}, our model parameters and inference time are more advantageous.
\begin{table}[!h]
	\centering
	\caption{Complexity comparison of TSM-Det.}
	\label{tablecomplex}
	\small
	\begin{tabular}{p{110pt}<{\centering}p{18pt}<{\centering}p{18pt}<{\centering}p{18pt}<{\centering}p{25pt}<{\centering}}
		\hline
		{Method}&Para. (M)&Max Batch&Infer (ms)&GFLOPs\cr
		\hline
		PointPillars\dag\cite{LANG2019POINTPILLARS} &4.83&7& 22.0&63.48\\
		PV-RCNN\dag\cite{SHI2020PVRCNN} &13.12&2& 75.9&93.11\\
		Voxel R-CNN\dag\cite{DENG2021VOXELRCNN} &35.77&7& 30.8&27.02\\
		X-Ray SECOND-Scaled\cite{gambashidze2024weaktostrong}&6.2&-&-&-\\
		{CPC-3Det\cite{PENG2024CPC3DET}}&10.29&8&26.1&32.73\\
		\hline
		\textbf{TSM-Det (Ours)}&\textbf{3.23}&\textbf{16}&\textbf{15.7}&\textbf{12.24}\\
		\hline
		\multicolumn{5}{p{180pt}}{\dag Reproduced. (Para.) Parameter.}
	\end{tabular}
	
\end{table}
\section{Discussion}
This paper implements single-neighborhood multi-scale feature approximation and constructs transferable class-aware knowledge between the teacher and the student. Additionally, this paper addresses the challenges posed by the center offset during training. In fact, all the work in this paper aims to reduce computational costs while approaching the performance of the teacher branch. As such, the performance of the teacher network will limit the capabilities. Furthermore, the extraction of discriminative features from sparse point clouds is crucial. For instance, some target objects possess fewer than 3 point clouds. We envision that employing the visual language model to enhance information and subsequently improve perception performance could be a viable solution. Note that TSM-Det has minimized the computational cost and can offer a fundamental solution for researchers with limited computing resources.
\section{Conclusion}
This paper presents the innovative detector TSM-Det, which includes the investigation of multi-scale feature approximation and transferable features.
Multi-scale neighborhood features are captured based on knowledge distillation and single-scale neighborhood points. 
Transferable class-aware statistics improve model performance by operating on classification and localization detection heads. 
In addition, this paper also considers the IoU misalignment brought by the center offset calculated by soft labels and predicted bounding boxes.


\begin{thebibliography}{0}
		
	    \bibitem{R1} G. Zamanakos, L. Tsochatzidis, A. Amanatiadis, I. Pratikakis, A comprehensive survey of LIDAR-based 3D object detection methods with deep learning for autonomous driving, \textit{Computers \& Graphics}, 2021, vol. 99, pp. 153-181.
		
		\bibitem{R2} T. Ku, S. Galanakis, B. Boom, R. C. Veltkamp, D. Bangera, S. Gangisetty, N. Stagakis, G. Arvanitis, K. Moustakas, \textit{Retraction notice to “SHREC 2021: 3D point cloud change detection for street scenes}, \textit{Computers \& Graphics}, 2024, vol. 125, pp. 104127.
		
		\bibitem{R3} Y. Gao, H. Yuan, T. Ku, et al, \textit{SHREC 2023: Point cloud change detection for city scenes}, \textit{Computers \& Graphics}, 2023, vol. 115, pp. 35-42.
		
		\bibitem{R4} Y. Lan, Y. Duan, Y. Shi, H. Huang, K. Xu, \textit{3DRM: Pair-wise relation module for 3D object detection}, \textit{Computers \& Graphics}, 2021, vol. 98, pp. 58-70.
		\bibitem{R5} J. Shu, S. Yu, X. Shu, J. Hu, \textit{SOA: Seed point offset attention for indoor 3D object detection in point clouds}, \textit{Computers \& Graphics}, 2024, vol. 123, pp. 0097-8493.
        
        \bibitem{DENG2021VOXELRCNN} J. Deng, S. Shi, P. Li, W. Zhou, Y. Zhang, and H. Li, \textit{Voxel rcnn: Towards high performance voxel-based 3d object detection}, in \textit{ Assoc. Advancement Artif. Intell.}(AAAI), 2020.
        
		\bibitem{VoxSet} C. He,  R. Li,  S. Li,  L. Zhang, Voxel Set Transformer: A Set-to-Set Approach to 3D Object Detection from Point Clouds, \textit{IEEE/CVF Conf. Comput. Vis. Pattern Recognit.}, 2022, pp.8417--8427.

        \bibitem{PENG2024CPC3DET} H. Peng and G. Tong, {Class-aware 3d detector from point clouds with partial knowledge diffusion and center-weighted iou}, \textit{IEEE Trans. Circuits Syst. Video Technol.}, vol. 34, no. 2, pp. 1043–1056, 2024.
        
		\bibitem{ZHENG2021CIASSD} W. Zheng,W. Tang, S. Chen, L. Jiang, and C.-W. Fu, {Cia-ssd: Confident iou-aware single-stage object detector from point cloud}, in \textit{ Assoc. Advancement Artif. Intell.} (AAAI), 2021.
		
		\bibitem{ZHENG2021SESSD} W. Zheng, W. Tang, L. Jiang, and C.-W. Fu, {Se-ssd: Self-ensembling single-stage object detector from point cloud}, in Proc. \textit{IEEE/CVF Conf. Comput. Vis. Pattern Recognit.} (CVPR), 2021, pp. 14489–14498.
		
		\bibitem{YANG2023GDMAE} H. Yang, T. He, J. Liu, H. Chen, B. Wu, B. Lin, X. He, and W. Ouyang, {Gd-mae: Generative decoder for mae pre-training on lidar point clouds}, in Proc. \textit{IEEE/CVF Conf. Comput. Vis. Pattern Recognit.} (CVPR), 2023.
		
		\bibitem{XIAO2023BSAODet} W. Xiao, Y. Peng, C. Liu, J. Gao, Y. Wu, and X. Li, {Balanced sample assignment and objective for single-model multi-class 3d object detection}, \textit{IEEE Trans. Circuits Syst. Video Technol.}, pp. 1–1, 2023.
		
		\bibitem{CHARLES2017POINTNET++} C. R. Qi, L. Yi, H. Su, and L. J. Guibas, {Pointnet++: Deep hierarchical feature learning on point sets in a metric space}, \textit{CoRR}, vol.abs/1706.02413, 2017.
		
		\bibitem{SHI2019POINTRCNN} S. Shi,  X. Wang, H. Li, {PointRCNN: 3D Object Proposal Generation and Detection From Point Cloud}, in Proc. \textit{IEEE/CVF Conf. Comput. Vis. Pattern Recognit.} (CVPR), 2019, pp. 770–779.
		
		\bibitem{YANG20203DSSD} Z. Yang, Y. Sun, S. Liu, and J. Jia, {3dssd: Point-based 3d single stage object detector}, in Proc. \textit{IEEE/CVF Conf. Comput. Vis. Pattern Recognit.} (CVPR), 2020, pp. 11 037–11 045.
		
		\bibitem{ZHANG2022NOT} Y. Zhang, Q. Hu, G. Xu, Y. Ma, J. Wan, and Y. Guo, {Not all points are equal: Learning highly efficient point-based detectors for 3d lidar point clouds}, in Proc. \textit{IEEE Conf. Comput. Vis. Pattern Recognit.} (CVPR), 2022, pp. 18953–18962.
		
		
		\bibitem{CHEN2022SASA} C. Chen, Z. Chen, J. Zhang, and D. Tao, {SASA: semantics-augmented set abstraction for point-based 3d object detection}, \textit{CoRR}, vol.abs/2201.01976, 2022.

        \bibitem{r2p1} Xia, Q., Ye, W., Wu, H., Zhao, S., Xing, L., Huang, X., Deng, J., Li, X., Wen, C. and Wang, C., Hinted: Hard instance enhanced detector with mixed-density feature fusion for sparsely-supervised 3D object detection. In Proc. \textit{IEEE/CVF Conf. Comput. Vis. Pattern Recognit.} (CVPR), 2024, pp. 15321-15330.
        
        \bibitem{r2p2} Park, J., Xu, C., Zhou, Y., Tomizuka, M. and Zhan, W., Detmatch: Two teachers are better than one for joint 2d and 3d semi-supervised object detection. In \textit{Eur. Conf. Comput. Vis.}, 2022, pp. 370-389. 
        
        \bibitem{r2p3} Liu, C., Gao, C., Liu, F., Li, P., Meng, D. and Gao, X., Hierarchical supervision and shuffle data augmentation for 3d semi-supervised object detection. In Proc. \textit{IEEE/CVF Conf. Comput. Vis. Pattern Recognit.} (CVPR), 2023, pp. 23819-23828.
        
        \bibitem{r2p4} Liu, C., Gao, C., Liu, F., Liu, J., Meng, D. and Gao, X.,  Ss3d: Sparsely-supervised 3d object detection from point cloud. In Proc. \textit{IEEE/CVF Conf. Comput. Vis. Pattern Recognit.} (CVPR), 2022, pp. 8428-8437.
		
		\bibitem{LANG2019POINTPILLARS} A. H. Lang, S. Vora, H. Caesar, L. Zhou, J. Yang, and O. Beijbom, {Pointpillars: Fast encoders for object detection from point clouds}, in Proc. \textit{IEEE/CVF Conf. Comput. Vis. Pattern Recognit.} (CVPR), 2019, pp. 12689–12697.
		
		\bibitem{HE2020SASSD} C. He, H. Zeng, J. Huang, X.-S. Hua, and L. Zhang, {Structure aware single-stage 3d object detection from point cloud}, in Proc. \textit{IEEE/CVF Conf. Comput. Vis. Pattern Recognit.} (CVPR), 2020, pp. 11870–11879.
		
		\bibitem{YAN2018SECOND} Y. Yan, Y. Mao,  B. Li, \textit{SECOND: Sparsely Embedded Convolutional Detection}, in \textit{Sensors}(AAAI), 2018.
		
		\bibitem{QIAN2022BANET} R. Qian, X. Lai, and X. Li, {Badet: Boundary-aware 3d object detection from point clouds}, \textit{Pattern Recognit.}, vol. 125, p. 108524, 2022.
		
		\bibitem{SHENG2021CT3D} H. Sheng, S. Cai, Y. Liu, B. Deng, J. Huang, X.-S. Hua, and M.-J. Zhao, {Improving 3d object detection with channel-wise transformer}, in \textit{IEEE/CVF Int. Conf. Comput. Vis.} (ICCV), 2021, pp. 2723–2732.
		
		\bibitem{HU2022PDV} J. S. K. Hu, T. Kuai, and S. L. Waslander, {Point density-aware voxels for lidar 3d object detection}, 2022.
		
		\bibitem{SHI2020PVRCNN} S. Shi, C. Guo, L. Jiang, Z. Wang, J. Shi, X. Wang, and H. Li, {Pvrcnn: Point-voxel feature set abstraction for 3d object detection}, in Proc. \textit{IEEE/CVF Conf. Comput. Vis. Pattern Recognit.} (CVPR), 2020, pp. 10526–10535.
		
		\bibitem{HE2022SVGANET} Q. He, Z. Wang, H. Zeng, Y. Zeng, S. Liu, and B. Zeng, {Svga-net: Sparse voxel-graph attention network for 3d object detection from point clouds}, \textit{CoRR}, vol. abs/2006.04043, 2020.
			
		\bibitem{LIANG2021RANGEIOUDET} Z. Liang, Z. Zhang, M. Zhang, X. Zhao, and S. Pu, {Rangeioudet: Range image based real-time 3d object detector optimized by intersection over union}, in Proc. \textit{IEEE/CVF Conf. Comput. Vis. Pattern Recognit.} (CVPR), 2021, pp. 7136–7145.
				
		\bibitem{MAHMOUD2022DVF} A. Mahmoud, J. S. Hu, and S. L. Waslander, {Dense voxel fusion for 3d object detection}, \textit{arXiv preprint} arXiv:2203.00871, 2022.
		\bibitem{MAO2021VOTR} J. Mao,  Y. Xue,  M. Niu,  H. Bai,  J. Feng,  X. Liang,  H. Xu,  C. Xu, Voxel Transformer for 3D Object Detection, \textit{IEEE/CVF Conf. Comput. Vis. Pattern Recognit.} (CVPR), 2021, pp. 3144-3153.
		
		\bibitem{LIU2024MSRMNet} X. Liu, L. Wang, MSRMNet: Multi-scale skip residual and multi-mixed features network for salient object detection, \textit{Neural Networks}, 2024, 173, pp. 106144. 
		
		\bibitem{DENG2021H23DRCNN} J. Deng, W. Zhou, Y. Zhang, and H. Li, {From multi-view to hollow-3d: Hallucinated hollow-3d r-cnn for 3d object detection}, \textit{IEEE Trans. Circuits Syst. Video Technol.}, vol. 31, no. 12, pp. 4722–4734, 2021.
		
		\bibitem{KITTI} A. Geiger,  P. Lenz,  C. Stiller,  R. Urtasun, Vision Meets Robotics: The KITTI Dataset, \textit{Sage Publications, Inc.}, 2013, vol.32, no. 11, pp.1231-1237.
		\bibitem{WAYMO} P. Sun,  H. Kretzschmar, X. Dotiwalla,  A. Chouard,  et al, Scalability in Perception for Autonomous Driving: Waymo Open Dataset, \textit{IEEE/CVF Conf. Comput. Vis. Pattern Recognit.}, 2020, pp.2443-2451.
		
		\bibitem{XU2021SPG} Q. Xu, Y. Zhou, W. Wang, C. R. Qi, and D. Anguelov, {Spg: Unsupervised domain adaptation for 3d object detection via semantic point generation}, in \textit{IEEE/CVF Int. Conf. Comput. Vis.} (ICCV), 2021, pp. 15426–15436.
		
		\bibitem{YANG2022EQPVRCNN} Z. Yang, L. Jiang, Y. Sun, B. Schiele, and J. Jia, {A unified query-based paradigm for point cloud understanding}, 2022.
		
		\bibitem{YIN2021CENTERNET} T. Yin, X. Zhou, and P. Kr¨ahenb¨uhl, {Center-based 3d object detection and tracking}, in Proc. \textit{IEEE/CVF Conf. Comput. Vis. Pattern Recognit.} (CVPR), 2021, pp. 11779–11788.
		
		
		\bibitem{gambashidze2024weaktostrong} A. Gambashidze, A. Dadukin, M. Golyadkin, M. Razzhivina, and I. Makarov, {Weak-to-strong 3d object detection with x-ray distillation}, 2024.
		
		
		\bibitem{MAO2021DSAPVRCNN} P. Bhattacharyya, C. Huang, and K. Czarnecki, {Sa-det3d: Self-attention based context-aware 3d object detection}, in \textit{IEEE/CVF Int. Conf. Comput. Vis. Workshops} (ICCVW), 2021, pp. 3022–3031.

        \bibitem{partA2} S. Shi, Z. Wang, J. Shi, X. Wang and H. Li, From Points to Parts: 3D Object Detection From Point Cloud With Part-Aware and Part-Aggregation Network, \textit{IEEE Trans. Pattern Anal. Mach. Intell.}, 2021, vol. 43, no. 8, pp. 2647-2664
		
		
		
		
			
		
	\end{thebibliography}
\end{document}